\documentclass[sigconf]{acmart}
\AtBeginDocument{%
  }

\usepackage{graphicx}
\usepackage{subcaption}
\usepackage{multirow}
\usepackage{tcolorbox}
\usepackage{balance}

\acmConference[KDD'26]{KDD'26,August 09–13, 2026, Jeju Island, Republic of Korea}{August 09–13, 2026}{Jeju Island, Republic of Korea}

\copyrightyear{2026}
\acmYear{2026}
\setcopyright{cc}
\setcctype{by}
\acmConference[KDD '26]{Proceedings of the 32nd ACM SIGKDD Conference on Knowledge Discovery and Data Mining V.2}{August 09--13, 2026}{Jeju Island, Republic of Korea}
\acmBooktitle{Proceedings of the 32nd ACM SIGKDD Conference on Knowledge Discovery and Data Mining V.2 (KDD '26), August 09--13, 2026, Jeju Island, Republic of Korea}
\acmDOI{10.1145/3770855.3818338}
\acmISBN{979-8-4007-2259-2/2026/08}

\begin{document}

\title{CEDAR: Controlled and Event-Driven Demand Forecasting via Residual Decomposition}

\author{Junjie Meng}
\orcid{0009-0006-5709-9963}
\affiliation{
  \institution{School of Artificial Intelligence and Data Science, University of Science and Technology of China}
  \city{Hefei}
  \country{China}
}
\email{mengjre@gmail.com}

\author{Ranxu Zhang}
\orcid{0009-0006-7222-2149}
\affiliation{
  \institution{School of Artificial Intelligence and Data Science, University of Science and Technology of China}
  \city{Hefei}
  \country{China}
}
\email{zkd_zrx@mail.ustc.edu.cn}

\author{Zi-an Zhang}
\orcid{0009-0008-8160-2156}
\affiliation{
  \institution{Alibaba Group}
  \city{Hangzhou}
  \country{China}
}
\email{zhangzian.zza@alibaba-inc.com}

\author{Shujun Liu}
\orcid{0009-0009-3449-6255}
\affiliation{
  \institution{Alibaba Group}
  \city{Hangzhou}
  \country{China}
}
\email{liushujun_uestc@163.com}

\author{Xiaoning Qi}
\orcid{0009-0001-1650-1577}
\affiliation{
  \institution{Alibaba Group}
  \city{Hangzhou}
  \country{China}
}
\email{xiaoning.qxn@alibaba-inc.com}

\author{Xiaozhou Xu}
\orcid{0009-0008-6290-9291}
\affiliation{
  \institution{Alibaba Group}
  \city{Hangzhou}
  \country{China}
}
\email{heixia.xxz@alibaba-inc.com}

\author{Yanyong Zhang}
\orcid{0000-0001-9046-798X}
\affiliation{
  \institution{School of Artificial Intelligence and Data Science, University of Science and Technology of China}
  \city{Hefei}
  \country{China}
}
\email{yanyongz@ustc.edu.cn}

\author{Hui Xiong}
\orcid{0000-0001-6016-6465}

\affiliation{%
  \institution{Thrust of Artificial Intelligence, The Hong Kong University of Science and Technology (Guangzhou)}
  \city{Guangzhou}
  \country{China}
}

\affiliation{%
  \institution{Department of Computer Science and Engineering, The Hong Kong University of Science and Technology}
  \city{Hong Kong SAR}
  \country{China}
}

\email{xionghui@ust.hk}

\author{Chao Wang}
\authornote{Chao Wang is the corresponding author.}
\orcid{0000-0001-7717-447X}
\affiliation{
  \institution{School of Artificial Intelligence and Data Science, University of Science and Technology of China}
  \city{Hefei}
  \country{China}
}
\email{wangchaoai@ustc.edu.cn}

\begin{abstract}
Forecasting in large-scale e-commerce marketplaces is increasingly required to support \emph{planning}: merchants need to evaluate sales outcomes under \emph{future action sequences} such as budget schedules, rather than passively predicting what happens next. 
However, most existing time series forecasting (TSF) approaches remain inherently passive. Even when incorporating operational decisions as auxiliary covariates, they typically optimize for correlation-based extrapolation under historical policies. This design suffers from {autoregressive inertia} and conflates {endogenous market evolution} with {decision-induced transitions}, leading to policy-insensitive rollouts and unreliable counterfactual analysis.
To bridge this gap, we propose \textbf{CEDAR} (\textbf{C}ontrolled and \textbf{E}vent-Driven \textbf{D}emand forecasting via \textbf{A}ction-aware \textbf{R}esidual decomposition), a two-stage framework for robust decision-conditioned simulation. 
In Stage~I, an \emph{Action-Interleaved Transformer} learns controllable action-conditioned state transitions for rollout under planned interventions. 
In Stage~II, a \emph{Residual Correction Module} leverages external event signals and LLM-assisted text representations to align noisy event descriptions with product context and correct event-driven deviations. 
Our study is enabled by a large-scale real-world dataset from Alibaba 1688, comprising approximately 32 million product trajectories with paired state--action sequences and aligned event signals. 
Extensive offline experiments and online controlled experiments in production demonstrate that CEDAR consistently improves simulation accuracy over strong TSF baselines and delivers practical gains for real-world budget planning.
\end{abstract}

\begin{CCSXML}
<ccs2012>
<concept>
<concept_id>10002951.10003227.10003351</concept_id>
<concept_desc>Information systems~Data mining</concept_desc>
<concept_significance>500</concept_significance>
</concept>
<concept>
<concept_id>10010405.10010481.10010487</concept_id>
<concept_desc>Applied computing~Forecasting</concept_desc>
<concept_significance>500</concept_significance>
</concept>
</ccs2012>
\end{CCSXML}

\ccsdesc[500]{Information systems~Data mining}
\ccsdesc[500]{Applied computing~Forecasting}

\keywords{Time Series Forecasting, Decision-Conditioned Simulation, Sales Forecasting}

\maketitle
\renewcommand{\shortauthors}{Junjie Meng et al.}

\section{Introduction}

\begin{figure*}[htbp]
    \centering
    \begin{subfigure}[b]{0.49\textwidth}
        \centering
        \includegraphics[width=\textwidth]{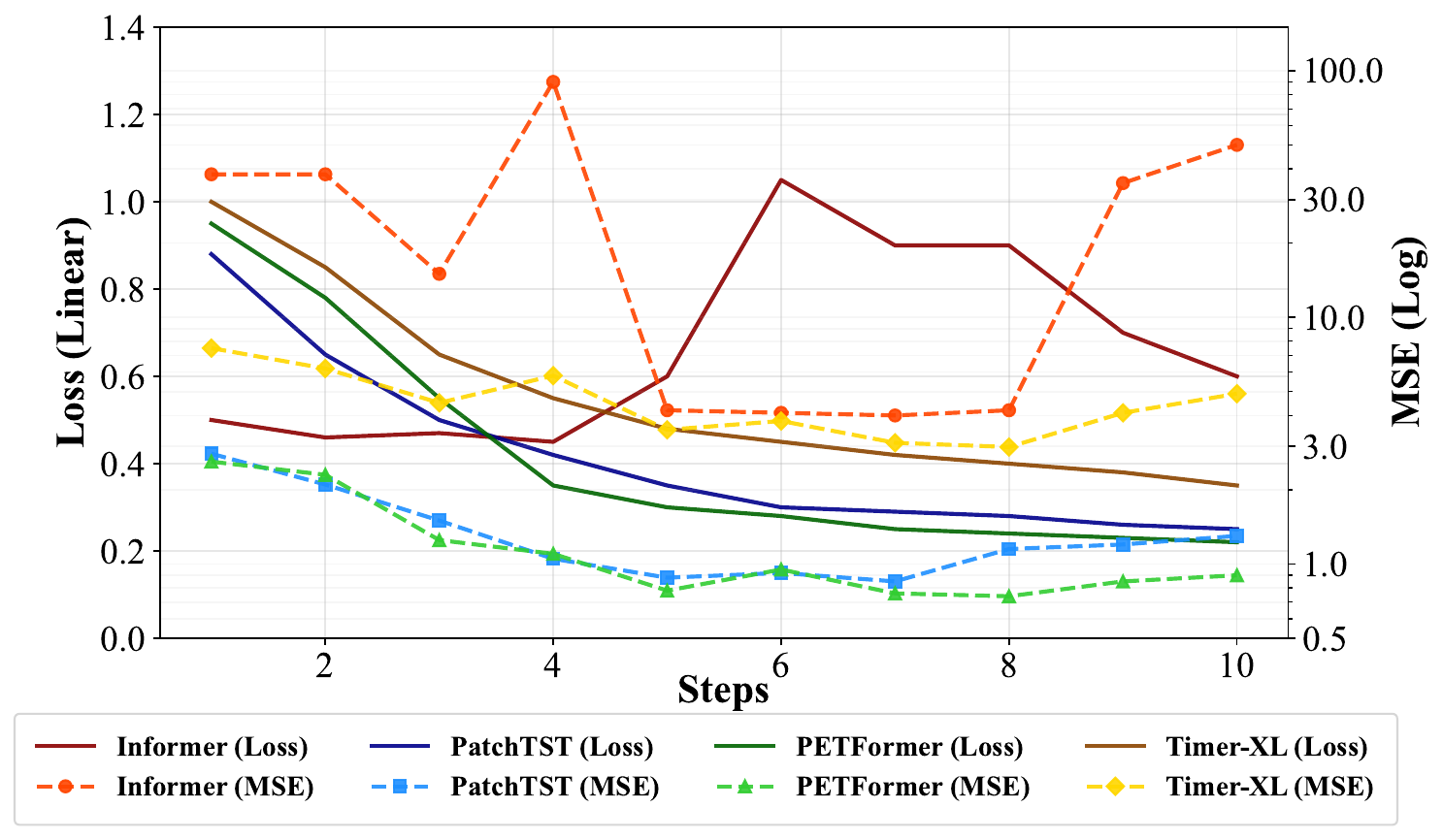}
        \caption{Training loss \& evaluation MSE visualization.}
        \label{fig:sub_a}
    \end{subfigure}
    \hfill
    \begin{subfigure}[b]{0.49\textwidth}
        \centering
        \includegraphics[width=\textwidth]{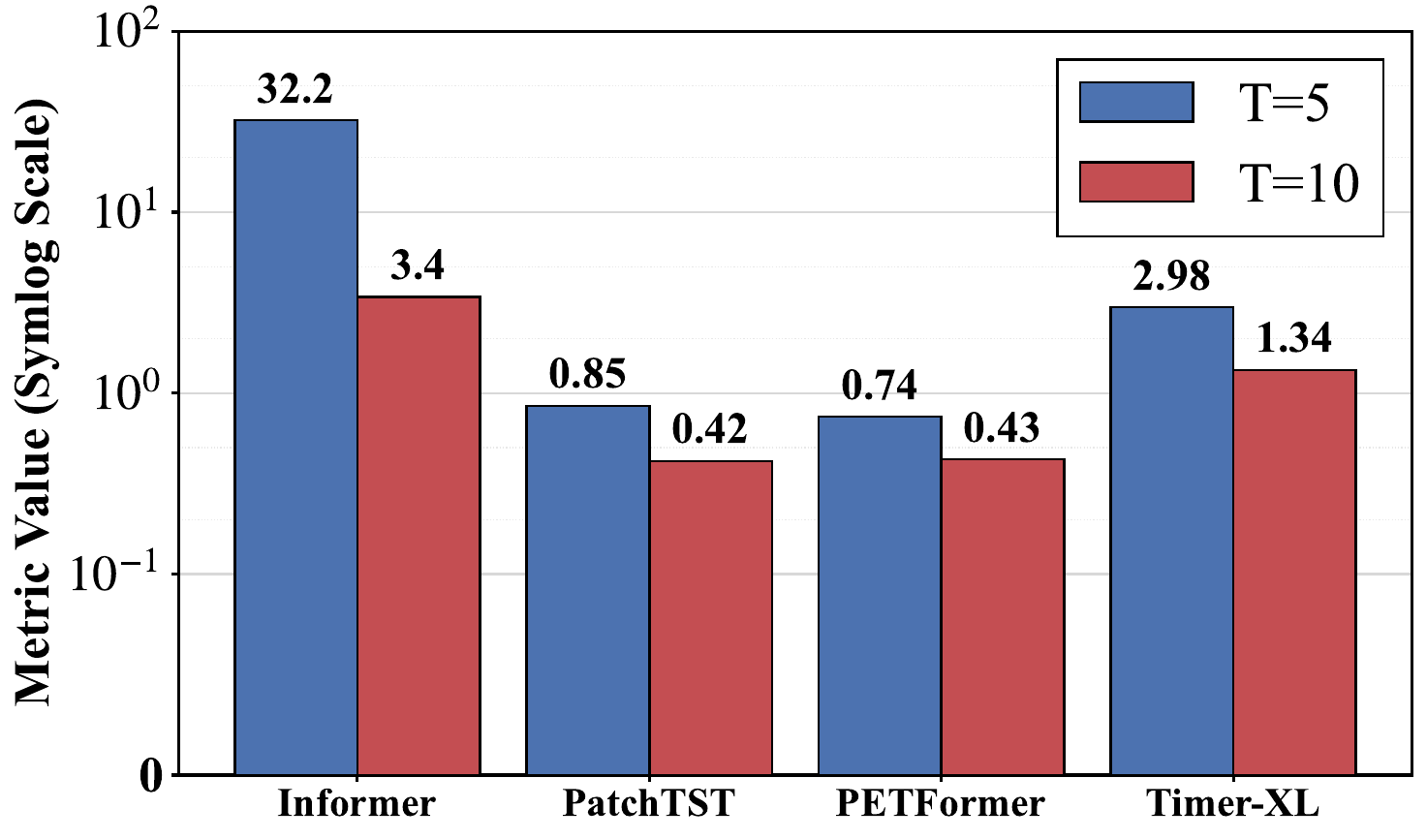}
        \caption{Comparison of multivariate time-series models. T represents the length of prediction.}
        \label{fig:sub_b}
    \end{subfigure}
    
    \caption{Performance comparison of four TSF baselines. All models are trained on the E-Com 15-Week dataset for 10 steps prediction. }
    \label{fig:main}
\end{figure*}

Time series forecasting (TSF)~\cite{wen2019robuststl, cleveland1990stl, goswamimoment, wang2018multilevel} is widely used in e-commerce to support inventory planning, pricing, and marketing optimization~\cite{das2024decoder, wen2020fastrobustSTL}. However, in large-scale marketplaces such as Alibaba 1688, forecasting is not merely about predicting the future; it is about evaluating \textit{planned interventions}. Merchants continuously act on the system---adjusting prices, launching promotions, and allocating advertising budgets---with the explicit goal of reshaping future demand trajectories. Consequently, the central question is not simply ``\textit{what will happen next?}'', but rather ``\textit{what would happen if I follow a particular budget schedule (and other actions) over the next weeks?}''.
This shifts the goal from passive forecasting to {decision-conditioned simulation}~\cite{aksugift, wilder2019melding}: given historical states/actions and a \emph{future action sequence}, the model should roll out the future demand trajectory under that plan.

Despite rapid progress in TSF architectures~\cite{nelson1998time, hu2024transforming}, most popular forecasters are still trained and evaluated in a \emph{passive} setting, ignoring the exogenous change. 
When directly applied to decision-conditioned rollout, such {state-only} TSF baselines (e.g., Informer~\cite{zhou2021informer}, PatchTST~\cite{nie2023patchtst}, PETFormer~\cite{lin2024petformer}, Timer-XL~\cite{liu2025timer}) exhibit strong \emph{autoregressive inertia}: they extrapolate along historical trends but cannot respond to counterfactual action schedules. 
As shown in Figure~\ref{fig:main}, these models perform poorly when action data is missing.

A natural remedy is to incorporate merchant operations into multivariate TSF models by treating actions (e.g., discounts, ad spend) as \emph{exogenous covariates}~\cite{salinas2020deepar, ansarichronos, ye2024survey, lim2021temporal}, and to further refine predictions via exogenous corrections such as RevPred~\cite{potapczynski2024effectively}. 
While this often improves short-horizon accuracy under historically observed policies, the covariate-fusion design is still limited for interventional simulation: it typically treats \emph{controllable interventions} indistinguishably from {passive context} by simply concatenating them with states. 
As a result, the model tends to capture correlations in the historical joint distribution~\cite{carta2018forecasting,qiu-etal-2017-alime}, entangling {endogenous market evolution} with {decision-induced transitions} and even mixing action effects with non-stationary exogenous shocks~\cite{10.1145/3774904.3792495, liu2026rethinking}.

Building a faithful decision-conditioned simulator therefore faces two pivotal challenges. 
The first is {controllable dynamics modeling}. To enable reliable rollouts, the model must learn how actions explicitly {drive} state transitions ($\mathbf{s}_{t-1} \xrightarrow{\mathbf{a}_{t}} \mathbf{s}_{t}$), rather than just fitting the joint distribution of states and actions.
The second is {exogenous disentanglement}. Real-world demand is frequently perturbed by non-stationary external factors (e.g., viral trends, holidays) that are not fully explainable by internal state-action history. 
If these exogenous shocks are not separated from action effects, the simulator will misattribute demand changes, leading to erroneous credit assignment and disastrous budget decisions. 
Therefore, we argue that robust what-if analysis in e-commerce requires an explicit, action-conditioned transition mechanism that is \emph{rollout-stable} under novel action sequences, together with a separate component to absorb non-stationary exogenous shocks. 

To this end, we propose \textbf{CEDAR} (\textbf{C}ontrolled and \textbf{E}vent-Driven \textbf{D}emand forecasting via \textbf{A}ction-aware \textbf{R}esidual decomposition), a novel framework designed for robust decision-conditioned simulation. 
Our study is enabled by a large-scale real-world dataset from Alibaba 1688, containing over 32 million product trajectories with paired \emph{state--action} sequences and aligned exogenous event signals, which also supports online evaluation in a production environment.
Departing from the monolithic covariate-fusion approach, CEDAR explicitly disentangles the sales generation process into two stages. 
In Stage~I, an \emph{Action-Interleaved Transformer (AIT)} models decision-conditioned state transitions by interleaving state and action tokens in the causal order $\mathbf{s}_{t-1}\rightarrow \mathbf{a}_t \rightarrow \mathbf{s}_t$, encouraging the backbone to capture how interventions drive subsequent state changes. 
In Stage~II, a \emph{Residual Correction Module} predicts the residual between the simulated trajectory and observations using external event signals, leveraging LLM-enhanced text understanding to align noisy, unstructured event descriptions with product contexts and correct event-driven deviations. 
This separation preserves controllability for planning while improving robustness to bursty shocks, enabling reliable multi-step simulation under alternative merchant action plans.

In summary, our contributions are:
\begin{itemize}
    \item We formulate merchant-facing sales forecasting as a {decision-conditioned simulation} problem, highlighting the limitations of passive TSF under sequential interventions.
    \item We propose \textbf{CEDAR}, a two-stage action-aware framework that (i) learns an explicit action-conditioned transition model via an Action-Interleaved Transformer, and (ii) corrects exogenous, non-stationary deviations with an event-driven residual module.
    \item We validate our approach on a massive industrial dataset of 32 million trajectories and conduct online A/B testing in a real production environment. Results demonstrate that CEDAR significantly outperforms state-of-the-art baselines in simulation accuracy and delivers substantial efficiency gains in real-world budget planning. We are also working to release a version of this dataset to the community.
\end{itemize}

\section{Related Work}
\subsection{General Time Series Forecasting and Covariate-Aware Modeling}

Time Series Forecasting (TSF)~\cite{zhou2025balm, WenRobustPeriod20} has evolved from statistical models and recurrent neural networks to Transformer-based architectures~\cite{wolff2024spade,hu2025multi} and large-scale foundation models. Informer introduces probabilistic sparse attention for efficient long-horizon modeling, while Autoformer~\cite{wu2021autoformer} and FEDformer~\cite{zhou2022fedformer} incorporate decomposition and frequency-domain learning to better capture complex temporal patterns. More recent methods such as PatchTST~\cite{nie2023patchtst} and iTransformer~\cite{liu2023itransformer} further refine representation learning through patch-wise tokenization and dimension inversion, achieving strong performance on multivariate forecasting benchmarks~\cite{qin2025scihorizon}.

Beyond pure autoregressive modeling, a parallel line of research incorporates rich covariates to improve multi-horizon forecasting~\cite{wang2021personalized, wang2023setrank}. Temporal Fusion Transformer (TFT)~\cite{punati2025temporal,lim2021temporal} represents a highly influential framework in this direction, integrating variable selection networks, gated residual connections, and attention mechanisms to adaptively fuse historical observations, known future inputs, and static features. Despite its empirical success, TFT~\cite{oliveira2024evaluating} fundamentally follows a covariate-conditioned forecasting paradigm, modeling $p(\mathbf{s}_{t+1} \mid \mathbf{s}_{\le t}, \mathbf{a}_{\le t})$ via feature-level fusion rather than explicitly learning action-conditioned state transitions. As a result, TFT primarily captures statistical correlations under historically observed policies, which limits its robustness under policy shifts and counterfactual simulation, where future action sequences deviate from training distributions.

More broadly, most multivariate TSF models~\cite{woo2022etsformer} treat controllable variables as auxiliary covariates, implicitly assuming invariant system dynamics~\cite{meng2026turningsemanticstopologyllmdriven}. This assumption is misaligned with decision-intensive environments such as e-commerce, where merchant actions actively reshape future trajectories. In contrast, our work formulates sales forecasting as a decision-conditioned simulation problem and explicitly models the transition operator $\mathbf{s}_{t+1}=f(\mathbf{s}_t,\mathbf{a}_{t+1})$. By interleaving state and action tokens in a unified Transformer, we encode the causal temporal ordering $\mathbf{s}_{t-1} \rightarrow \mathbf{a}_t \rightarrow \mathbf{s}_t$, enabling robust learning of action-conditioned dynamics and stable long-horizon rollout under novel intervention strategies.

\subsection{Offline Reinforcement Learning via Sequence Modeling}

Recent advances in offline reinforcement learning~\cite{levine2020offline, nakamoto2023cal,huang2024efficient, kumar2020conservative, wang2021variable} have reformulated policy learning and planning as a conditional sequence modeling problem. Decision Transformer (DT)~\cite{chen2021decisiontransformer} pioneered this paradigm by conditioning autoregressive transformers on desired return-to-go, demonstrating that generic sequence models can achieve competitive control performance without explicit dynamic programming. Trajectory Transformer~\cite{janner2021offline} further discretized continuous trajectories into token sequences and enabled long-horizon planning via beam search. Subsequent extensions improved data efficiency and deployment flexibility, including Online Decision Transformer~\cite{zheng2022online}, Q-learning Decision Transformer (QDT)~\cite{yamagata2023q}, and Value-Guided Decision Transformer (VDT)~\cite{zheng2025valueguided}, which introduced critic guidance, advantage weighting, and value regularization to stabilize learning and mitigate compounding errors.

Despite their success in control-centric benchmarks, these approaches are not directly applicable to merchant-facing sales forecasting and simulation. In typical e-commerce scenarios, the primary quantity of interest is the future sales trajectory itself, which simultaneously plays the role of system state and optimization objective. More fundamentally, existing DT-style models are designed to infer an implicit policy that maximizes long-term return under a fixed environment, rather than to explicitly learn a controllable state transition mechanism. In merchant budget planning, however, the core requirement is not to discover a single optimal policy, but to evaluate and compare multiple candidate strategies through reliable \emph{what-if} simulation. This necessitates disentangling endogenous market dynamics from action-induced transitions~\cite{10.1145/3774904.3792495,liu2026rethinking}, and explicitly modeling how alternative interventions reshape future trajectories. Recent explorations in causal generative modeling, such as \textit{DoFlow}~\cite{wu2025doflow}, have highlighted the necessity of incorporating causal flows for robust interventional and counterfactual time-series prediction~\cite{rafetseder2013counterfactual}, ensuring that simulated trajectories remain physically and logically consistent under distribution shifts.

In contrast to policy-centric sequence modeling, our work focuses on learning an action-conditioned state transition operator for stable multi-step rollout. By separating controllable effects from latent external fluctuations, CEDAR enables robust trajectory simulation under diverse intervention plans, providing a principled foundation for decision-aware forecasting and strategic exploration in highly non-stationary e-commerce environments.

\begin{figure*}[t]
    \centering
    \includegraphics[width=\linewidth]{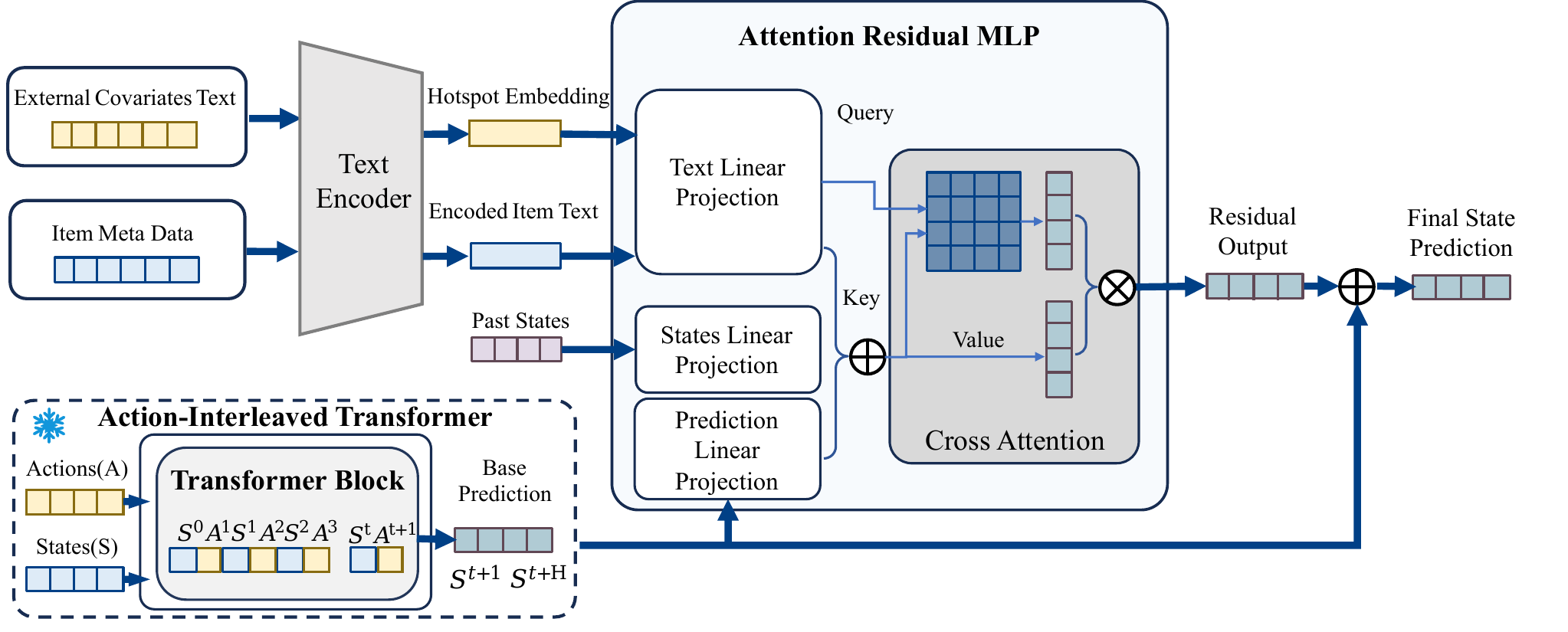}
    \caption{Framework of CEDAR. The AIT module is trained in the first stage, and frozen in the second stage.}
    \label{fig:placeholder}
\end{figure*}

\section{Method}

We formalize merchant-facing sales forecasting as a \emph{decision-conditioned simulation} problem. At each discrete time step $t$, the system is characterized by a product state vector $\mathbf{s}_t \in \mathbb{R}^{d_s}$ that captures endogenous signals such as impressions, clicks, favorites, and purchases, together with a merchant action vector $\mathbf{a}_t \in \mathbb{R}^{d_a}$ encoding controllable interventions including discount strategies and marketing expenditures. Both states and actions are aggregated over a fixed temporal granularity (e.g., weekly) in the E-Com 15-week dataset.

Given historical observations $\{\mathbf{s}_{1:T}, \mathbf{a}_{1:T}\}$ and a planned future action sequence $\mathbf{a}_{T+1:T+H}$, our objective is to simulate the counterfactual evolution of product trajectories $\mathbf{s}_{T+1:T+H}$ under alternative merchant policies. This formulation enables forward rollout of system dynamics conditioned on hypothetical budget allocation strategies, thereby supporting downstream tasks such as operational planning, strategy evaluation, and optimal budget scheduling. Unlike conventional TSF settings that focus on passive extrapolation, our goal is to learn a controllable state transition model that enables stable long-horizon simulation and reliable what-if analysis.

\subsection{Decision-Conditioned Transition Modeling}

Traditional multivariate TSF formulations typically aim to learn a conditional distribution $p(\mathbf{s}_{t+1} \mid \mathbf{s}_{\le t}, \mathbf{a}_{\le t})$, where merchant actions are treated as auxiliary covariates. While effective for short-term prediction under historically observed policies, this paradigm fundamentally conflates endogenous system evolution and decision-induced dynamics, resulting in models that primarily capture correlations rather than learning explicit action-conditioned transition mechanisms. 

In contrast, decision-conditioned simulation requires modeling a controlled dynamical system governed by a transition operator
\begin{equation}
\mathcal{T}: (\mathbf{s}_t, \mathbf{a}_{t+1}) \mapsto \mathbf{s}_{t+1},
\end{equation}
which describes how merchant interventions actively shape future trajectories. This formulation aligns with the core requirement of budget planning and strategy exploration, where future actions are intentionally optimized and may deviate significantly from historical distributions. Learning such an explicit transition operator enables faithful rollout under hypothetical action sequences and mitigates the extrapolation failures commonly observed in covariate-based TSF models. While some approaches incorporate external context, they typically fail to disentangle endogenous system dynamics from exogenous market shocks, leading to confounded state representations.

Motivated by this perspective, we design the Action-Interleaved Transformer (AIT), which treats both states and actions as first-class tokens and explicitly encodes their causal temporal ordering, thereby introducing a structural inductive bias toward action-conditioned transition modeling.

\subsection{Stage I: Action-Interleaved Transformer}

The first stage of CEDAR focuses on learning action-conditioned state transition dynamics. Instead of concatenating actions as exogenous features, we model both states and actions as interleaved tokens in a unified temporal sequence. Concretely, for each time step $t$, we construct a token ordering
\begin{equation}
 \mathbf{s}_{t-1} \rightarrow \mathbf{a}_t \rightarrow \mathbf{s}_t,
\end{equation}
which reflects the natural causal structure of merchant operations: historical system states inform merchant decisions, and these decisions subsequently drive state transitions.

Each state token and action token is first projected into a shared embedding space through separate linear encoders, preserving their semantic roles while enabling joint attention. A causal Transformer backbone is then applied over the interleaved sequence, ensuring that predictions at time $t+1$ depend only on past and present information. Formally, the model learns a parametric transition function
\begin{equation}
  \hat{\mathbf{s}}_{t+1} = f_\theta(\mathbf{s}_{\le t}, \mathbf{a}_{\le t+1}),
\end{equation}
where $f_\theta$ is implemented by the Action-Interleaved Transformer.

The interleaving design imposes a structured attention pattern that explicitly aligns merchant actions with subsequent state transitions, enabling the model to learn directed influence pathways such as $\mathbf{a}_t \rightarrow \mathbf{s}_t$. Compared to covariate-based architectures, this design encourages the Transformer to focus on controllable dynamics, improving stability and generalization when simulating under novel policies.

Furthermore, product state vectors exhibit strong internal structure, including funnel-like dependencies from exposure to conversion. The self-attention mechanism within AIT naturally captures such hierarchical interactions while modeling their modulation by merchant actions, enabling fine-grained transition learning across heterogeneous behavioral signals.

AIT is trained using a one-step prediction loss over observed trajectories:
\begin{equation}
\mathcal{L}_{\text{AIT}} = \sum_{t} \| \hat{\mathbf{s}}_{t+1} - \mathbf{s}_{t+1} \|_2^2.
\end{equation}

\subsection{Stage II: Residual Correction with External Signals}

While AIT captures endogenous dynamics and the direct effects of merchant actions, real-world sales trajectories are also influenced by latent, time-varying exogenous factors, including breaking news, social media trends, seasonal effects, and macroeconomic shocks. These influences introduce non-stationary perturbations that are difficult to infer solely from historical state-action trajectories, particularly in highly volatile markets.

To account for such effects without compromising the controllability and stability of the learned transition operator, we introduce a Residual Correction Module that explicitly models deviations induced by external factors. Specifically, we decompose the system evolution as
\begin{equation}
\mathbf{s}_{t+1} = f_\theta(\mathbf{s}_{\le t}, \mathbf{a}_{\le t+1}) + \epsilon_t,
\end{equation}
where $f_\theta$ captures controllable dynamics and $\epsilon_t$ represents latent external perturbations. Stage I models the former, while Stage II estimates the latter.

\paragraph{External Signal Construction.}
We leverage a large language model (LLM) to extract structured representations of exogenous market signals~\cite{wang2026face, meng2026turningsemanticstopologyllmdriven}. For each time window $t$, we collect news from the previous week and prompt the LLM to extract product-level keywords indicative of potential demand surges. Simultaneously, major holidays and seasonal events occurring in the current window are appended as additional signals. Rather than treating these cues as isolated tags, we instruct the LLM to synthesize them into a coherent natural language description, which is subsequently encoded by a pre-trained text encoder to produce a dense \emph{hotspot embedding} $\mathbf{h}_t$ that captures the semantic context of external influences. Details of hotspot embedding generation can be found in Appendix~\ref{app.hostspot}.

\paragraph{Residual Prediction.}
As shown in Figure~\ref{fig:placeholder}, we construct \emph{item status embedding} through concatenating the embeddings of item titles and the tags, the predicted next state $\hat{\mathbf{s}}_{t+1}$ and recent historical states $\mathbf{s}_{t-k:t}$.
The hotspot embedding $\mathbf{h}_t$ is then combined with the item status embedding via a cross-attention module, enabling dynamic alignment between external events and product-level temporal patterns. The resulting representation is passed through a multi-layer perceptron to estimate the residual correction:
\begin{equation}
    \Delta \mathbf{s}_{t+1} = g_\phi(\mathbf{h}_t, \hat{\mathbf{s}}_{t+1}, \mathbf{s}_{t-k:t}).
\end{equation}

 The loss of Stage-II training is calculated through: 
\begin{equation}
\mathcal{L}_{\text{RC}} = \sum_{t} \| {\mathbf{s}}_{t+1} - \hat{\mathbf{s}}_{t+1} - \Delta \mathbf{s}_{t+1}\|_2^2.
\end{equation}

The final simulated state is then obtained as
\begin{equation}
    \tilde{\mathbf{s}}_{t+1} = \hat{\mathbf{s}}_{t+1} + \Delta \mathbf{s}_{t+1}.
\end{equation}

\subsection{Discussion}

\subsubsection{Why exclude temporal signals}

During our empirical investigation, we also explored explicitly modeling periodic temporal signals and injecting them as additional perturbation embeddings into the predictive sequence. However, experimental results indicate that such temporal embeddings are difficult to align with the semantic representations of bursty external events extracted from text, leading to ineffective fusion. Moreover, explicitly modeling standalone temporal embeddings substantially increases the learning complexity of the network, resulting in degraded predictive performance.

We hypothesize that, \textit{in e-commerce settings, the primary performance gains attributed to temporal signals largely stem from holiday-driven demand fluctuations}. Since major holidays and seasonal events are already explicitly incorporated into the Residual Correction Module through LLM-based external signal construction, introducing timestamp embeddings as an additional perturbation provides limited marginal benefit. Consequently, we do not include explicit timestamp embeddings in CEDAR.

\subsubsection{The design of Residual Correction} 
While prior work~\cite{potapczynski2024effectively} similarly utilized a residual module to refine predictions from the backbone model, it relied on simple MLPs to process external covariates independently. This approach treats all covariates uniformly, thereby overlooking the semantic alignment between external shocks and specific items. For instance, the demand for seasonal items, such as \textit{Christmas hats}, is unlikely to surge during the \textit{Chinese New Year}, despite the presence of a significant festival event. In contrast, CEDAR employs a cross-attention mechanism. Specifically, the encoded external shock embeddings serve as queries to explicitly explore and model the relevance between global events and item-specific dynamics. \textbf{This mechanism enables CEDAR to dynamically align external impacts with internal product states, thereby significantly enhancing the model's effectiveness in capturing complex dependencies.}

\subsubsection{Training and Inference Pipeline}

\paragraph{Training Phase} We adopt a decoupled two-stage training paradigm. This design choice is primarily motivated by the need to isolate controllable system dynamics from stochastic external perturbations. By separating the learning process, we ensure that the action-conditioned transition operator remains stable under significant policy shifts, while the residual correction module can flexibly adapt to non-stationary market conditions. This decoupling prevents the model from over-relying on exogenous correlations, which empirically leads to substantial improvements in long-horizon rollout accuracy and enhances the robustness of "what-if" strategy simulations.

\paragraph{Inference Phase} During the iterative simulation process, the state for each successive time step is generated through a sequential execution of the model components, where the Action-Interleaved Transformer first generates an initial state estimate based on the historical trajectory and planned interventions, which is then refined by the Residual Correction Module to account for adjustments necessitated by latent external signals. The final predicted state for the time step is obtained as the sum of these two outputs; to perform multi-step forecasting, this result is appended to the historical sequence and fed back into the model in an auto-regressive manner, enabling the stable simulation of future product trajectories over an extended horizon.

\begin{table*}[t]
\centering
\caption{Performance comparison of CEDAR and other baselines.
The \textit{next 5} setting predicts the next 5 weeks given the previous 10 weeks,
while the \textit{next 10} setting predicts the next 10 weeks given the previous 5 weeks.
All offline results are reported as mean $\pm$ standard deviation over five independent runs.}
\label{tab:compare1}
\setlength{\tabcolsep}{4.5pt}
\renewcommand{\arraystretch}{1.08}
\begin{tabular}{llcccccc}
\toprule
\textbf{Metric} & \textbf{Setting}
& \textbf{CEDAR}
& \textbf{Informer}
& \textbf{TFT}
& \textbf{PatchTST}
& \textbf{PETFormer}
& \textbf{Timer-XL} \\
\midrule

\multirow{2}{*}{MSE $\downarrow$}
& \textit{next 10}
& \textbf{0.414$\pm$0.015}
& 32.2$\pm$21
& 0.722$\pm$0.024
& 0.849$\pm$0.026
& 0.743$\pm$0.021
& 2.98$\pm$0.14 \\
& \textit{next 5}
& \textbf{0.182$\pm$0.006}
& 3.44$\pm$1.8
& 0.572$\pm$0.015
& 0.424$\pm$0.011
& 0.434$\pm$0.013
& 1.34$\pm$0.05 \\

\midrule

\multirow{2}{*}{MAE $\downarrow$}
& \textit{next 10}
& \textbf{0.132$\pm$0.002}
& 3.23$\pm$2.1
& 0.194$\pm$0.007
& 0.201$\pm$0.006
& 0.192$\pm$0.006
& 0.627$\pm$0.035 \\
& \textit{next 5}
& \textbf{0.0603$\pm$0.001}
& 0.730$\pm$0.35
& 0.175$\pm$0.005
& 0.129$\pm$0.004
& 0.139$\pm$0.004
& 0.472$\pm$0.018 \\

\midrule

\multirow{2}{*}{NMSE $\downarrow$}
& \textit{next 10}
& \textbf{0.189$\pm$0.004}
& 14.7$\pm$9.2
& 0.337$\pm$0.012
& 0.387$\pm$0.013
& 0.339$\pm$0.011
& 1.36$\pm$0.08 \\
& \textit{next 5}
& \textbf{0.0830$\pm$0.001}
& 1.57$\pm$0.80
& 0.267$\pm$0.009
& 0.191$\pm$0.006
& 0.198$\pm$0.007
& 0.612$\pm$0.022 \\

\bottomrule
\end{tabular}
\end{table*}

\section{Experiments}

In this section, we conduct extensive experiments to systematically evaluate the effectiveness of the proposed CEDAR framework, including both the \emph{Action-Interleaved Transformer (AIT)} and the \emph{Residual Correction Module}. Our evaluation is designed to answer the following key research questions:

\begin{itemize}
\item \textbf{RQ1}: How does \textbf{CEDAR} perform compared to state-of-the-art time series forecasting baselines in decision-conditioned sales simulation?

\item \textbf{RQ2}: To what extent does explicitly modeling merchant actions as a dedicated modality improve forecasting accuracy and rollout stability?

\item \textbf{RQ3}: Can the proposed \emph{Residual Correction Module} effectively capture latent external impacts and further refine simulated trajectories?

\item \textbf{RQ4}: What is the real-world impact of deploying \textbf{CEDAR} in online budget planning scenarios on merchant engagement and platform performance?

\end{itemize}

\subsection{Evaluation Setup}
\subsubsection{Dataset}

To support large-scale training and evaluation, we construct a comprehensive e-commerce dataset from Alibaba 1688, covering approximately 32 million product trajectories spanning the years 2024 and 2025. Each trajectory consists of a sequence of \emph{product states} and \emph{merchant actions} aggregated over rolling 7-day windows.

Specifically, the state vector includes nine core indicators: 7-day aggregated impressions, page views, favorites, add-to-cart events, number of buyers, gross merchandise volume (GMV), advertisement impressions, advertisement clicks, and the number of customer inquiries. The action vector contains two controllable variables: the 7-day average marketing discount (defined as the ratio between the discounted price and the original price) and the total advertising expenditure during the same period.

 We segment each trajectory into overlapping windows of 15 consecutive weeks. A new window is sampled every 15 weeks, while remaining weeks within each year are backward-sampled to construct additional windows. This process yields approximately 32 million training samples. To fully evaluate the model's ability on unseen external shocks, we reserve the final window of 2025 as the test set and use all preceding windows for training, because random sampling may cause shortcut learning on already observed external shocks.

Each product in the dataset is annotated with a three-level category taxonomy, consisting of a primary category, secondary category, and fine-grained subcategory. At the finest granularity, the taxonomy contains 8,942 distinct subcategories. Due to the substantial heterogeneity in scale across product categories, we perform category-wise normalization. Specifically, for each subcategory, we compute the mean and standard deviation of all nine state variables and two action variables, and apply z-score normalization within each subcategory to mitigate scale disparities and stabilize model training. 

To account for abrupt external shocks and bursty demand fluctuations, we further incorporate exogenous event signals derived from both on-platform trending search topics and off-platform public hotspots, along with a curated list of 26 major holidays. These signals are used to support the modeling of latent external factors, with representative examples provided in the Appendix.

As the largest domestic B2B wholesale platform in China, Alibaba 1688 provides a uniquely rich environment for studying decision-conditioned forecasting and budget planning. The resulting \textbf{E-Comm 15-Week} dataset represents one of the largest real-world benchmarks for action-aware sales simulation, and we are actively working toward releasing a partially anonymized version to facilitate future research.

\subsubsection{Baselines}

We compare \textbf{CEDAR} against a diverse set of representative time series forecasting baselines, covering both classical architectures and recent foundation models. The selected methods span convolutional, attention-based, patch-wise, and large-scale pretrained paradigms, providing a comprehensive evaluation across different modeling principles. Specifically, we consider:

\begin{itemize}
\item \textbf{Informer}~\cite{zhou2021informer} is a canonical Transformer-based forecasting model that introduces probabilistic sparse attention to efficiently model long-range temporal dependencies. 

\item \textbf{TFT}~\cite{lim2021temporal} integrates exogenous covariates by variable selection networks, gated residual connections and attention mechanisms. 

\item \textbf{PatchTST}~\cite{nie2023patchtst} reformulates time series forecasting by segmenting the input sequence into local patches and applying channel-independent Transformer encoders. 

\item \textbf{PETFormer}~\cite{lin2024petformer} integrates periodicity-aware encoding and efficient temporal attention mechanisms to explicitly capture seasonal structures and long-term dependencies.

\item \textbf{Timer-XL}~\cite{liu2025timer} is a large-scale pretrained time series foundation model trained on heterogeneous real-world corpora.

\end{itemize}

For a fair comparison, all baselines are trained under identical data splits, input horizons, and prediction windows. All offline results are reported as the mean and standard deviation over five independent runs with different random seeds. This protocol allows us to evaluate whether the observed improvements are robust to optimization randomness rather than arising from a single favorable initialization. For models that support exogenous covariates, merchant actions are concatenated with the state variables as additional input channels. This setup reflects the prevailing practice in multivariate TSF, where actions are treated as auxiliary contextual features. In contrast, CEDAR explicitly models merchant actions as a first-class modality and interleaves them with state transitions, enabling direct learning of decision-conditioned dynamics and stable multi-step rollouts under alternative action plans. A detailed wall-clock cost analysis is provided in Appendix~\ref{app:time_cost}.

\begin{table}[t]
\centering
\caption{Ablation study results using MSE and MAE metrics. One-stage training refers to jointly training the Action-Interleaved Transformer and the Residual Correction Module.}
\label{tab:ablation_study}
\begin{tabular}{lcccc}
\toprule
\multirow{2}{*}{Ablation Variants} & \multicolumn{2}{c}{MSE $\downarrow$} & \multicolumn{2}{c}{MAE $\downarrow$} \\
\cmidrule(lr){2-3} \cmidrule(lr){4-5}
& \textit{next 10} & \textit{next 5} & \textit{next 10} & \textit{next 5} \\
\midrule
Full Model            & \textbf{0.414} & \textbf{0.182} & \textbf{0.132} & 0.0603 \\
w/o Recent states       & 0.443 & 0.209 & 0.137 & 0.0634 \\
w/o Product metadata & 0.466 & 0.227 & 0.143 & 0.0654 \\
w/o AIT Prediction          & 0.499 & 0.264 & 0.181 & 0.0697 \\
w/o Holiday keywords    & 0.451 & 0.213 & 0.169 & 0.0651 \\
w/o News keywords       & 0.417 & 0.188 & 0.141 & \textbf{0.0601} \\
w/o All(AIT only)          & 0.489 & 0.252 & 0.177 & 0.0691 \\
Temporal shuffle & 0.527 & 0.274 & 0.194 & 0.0712 \\
One stage         & 0.471 & 0.231 & 0.158 & 0.0674 \\
\bottomrule
\end{tabular}
\end{table}

\subsubsection{Evaluation Metrics \& Implementation Details}

We adopt three widely used time series forecasting metrics, including Mean Squared Error (MSE), Mean Absolute Error (MAE), and Normalized Mean Squared Error (NMSE). The NMSE is computed as $NMSE = \frac{MSE}{Mean(y^2)}$. To better align with real-world merchant requirements, we consider two evaluation settings: (i) forecasting the next 10 weeks given the past 5 weeks of observations, and (ii) forecasting the next 5 weeks given the past 10 weeks of observations. 

All experiments are conducted on a cluster equipped with 4 NVIDIA H20 GPUs for both training and inference. The hidden dimension of all models is uniformly set to 256. For Transformer-based models, including {CEDAR}, we configure the number of attention heads and layers as $n_{\text{head}}=4$ and $n_{\text{layer}}=5$, respectively. 

Given that the majority of textual data in our dataset is in Chinese, we adopt the BGE-zh-v1.5 model as our text encoder, with an embedding dimension of 1024. For hotspot keyword extraction and sentence organization, we employ the Qwen-Plus model.

\begin{figure}
    \centering
    \includegraphics[width=0.49\textwidth]{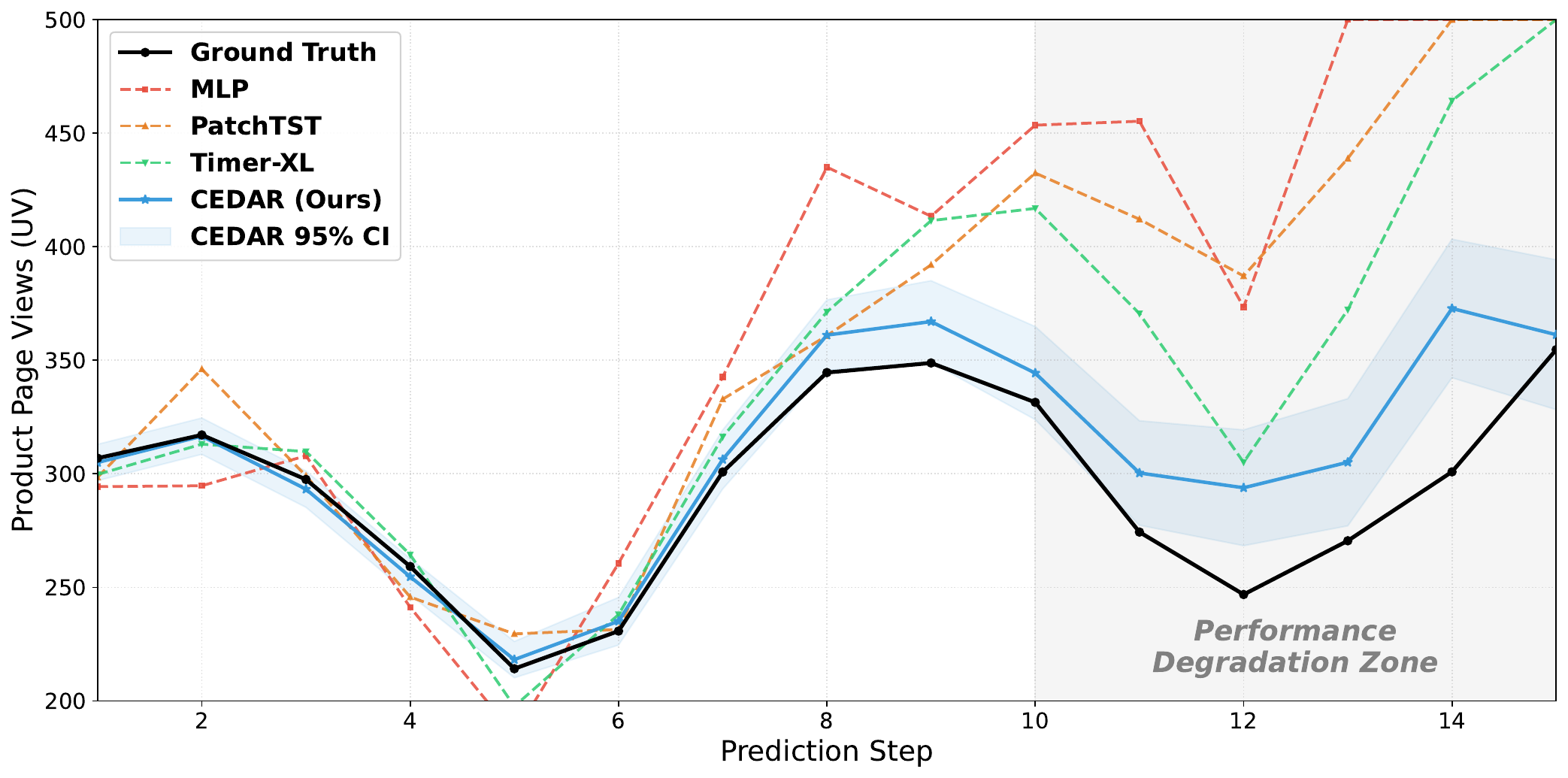}
    \caption{Visualization of predicted views trajectories of different baselines.}
    \label{fig:long_horizon}
\end{figure}

\subsection{Validation of CEDAR (RQ1)}

Table~\ref{tab:compare1} reports the quantitative comparison between CEDAR and a diverse set of representative TSF baselines, including Informer, PatchTST, PETFormer, and Timer-XL, under different forecasting horizons. We observe that CEDAR consistently achieves the best performance across all metrics and horizons, demonstrating substantial improvements over existing methods.

Specifically, at horizon \textit{next 5}, CEDAR attains an MSE of $0.182$, significantly outperforming the strongest baseline PatchTST $0.424$ and PETFormer $0.434$, corresponding to relative improvements of $57.1\%$ and $58.1\%$, respectively. Similar trends are observed under NMSE, where CEDAR reduces the error to $0.083$, yielding more than $56\%$ improvement over the best baseline. For the \textit{next 10} horizon, CEDAR also achieves the lowest MSE $0.414$ and NMSE $0.189$, consistently outperforming all competing approaches.

Notably, classical TSF models such as Informer suffer from severe performance degradation, with MSE exceeding $30$ at \textit{next 10}. This phenomenon highlights the limitation of decision-unaware forecasting models, which fail to capture the strong and non-stationary effects induced by merchant actions (e.g., pricing and traffic investments). In contrast, CEDAR explicitly models the interaction between system states and merchant decisions, enabling robust and accurate simulation under dynamic, intervention-driven environments. Besides, the ablation results in Table ~\ref{tab:ablation_study} also prove the effectiveness of two-stage training. 

To further evaluate CEDAR's performance on long-horizon prediction, we conduct additional long-horizon forecasting experiments, with detailed results reported in Appendix~\ref{app:Long-horizon}.

Overall, these results verify the superiority and robustness of CEDAR in decision-conditioned sales forecasting, particularly in scenarios characterized by strong action-driven distribution shifts.

\subsection{Action Modality Ablation (RQ2)}
To evaluate the impact of explicitly modeling merchant actions as a dedicated modality, we conduct a targeted analysis focusing on traffic forecasting, where advertising expenditure exhibits the strongest correlation with future dynamics. We compare CEDAR against multiple baselines that incorporate actions as auxiliary covariates, including PatchTST, Timer-XL, as well as an additional multilayer perceptron (MLP) baseline that directly concatenates historical states, product metadata, and action variables as input features.

Figure~\ref{fig:long_horizon} illustrates the predicted traffic trajectories for a representative product, where the merchant launches two advertising campaigns starting at weeks 5 and 12, respectively. Notably, baselines that treat actions as exogenous covariates exhibit limited sensitivity to intervention signals, particularly during the early phase of training. In the first 10 weeks, these models tend to extrapolate local trends from historical traffic peaks, with predicted local maxima typically lagging behind previously observed high-traffic points. This behavior reflects a strong reliance on autoregressive correlations, causing action effects to be largely diluted by dominant state dynamics.

In contrast, CEDAR, by explicitly interleaving actions with state transitions, demonstrates substantially improved responsiveness to marketing interventions. In particular, CEDAR accurately captures the decline in future traffic following the reduction of advertising expenditure, correctly anticipating the downward trend rather than merely propagating historical patterns. This ability enables stable and realistic multi-step rollouts under dynamically changing action plans, which is critical for budget planning and strategy exploration.

These findings indicate that modeling merchant actions as a first-class modality introduces a crucial structural inductive bias, allowing the model to disentangle endogenous temporal dynamics from decision-induced transitions. As a result, CEDAR achieves superior forecasting accuracy and significantly enhanced rollout stability in counterfactual scenarios involving policy shifts.

\subsection{External Shock Controls (RQ3)}

To assess whether the proposed \emph{Residual Correction Module} effectively captures latent external impacts and refines simulated trajectories, we conduct a dedicated ablation study, with quantitative results reported in Table~\ref{tab:ablation_study}. These variants remove or perturb different input sources used by the Residual Correction Module. In addition, we include a temporal-shuffle variant, where holiday and hotspot signals are randomly reassigned to different dates. This variant performs worse than the aligned-event setting and even degrades relative to AIT-only in next-5 MSE, indicating that CEDAR benefits from temporally meaningful event-demand alignment rather than merely using event embeddings as generic auxiliary features.

Overall, we observe that incorporating the Residual Correction Module yields a modest improvement in MSE but leads to a substantial reduction in MAE. This discrepancy is expected, as external shocks often manifest as abrupt, localized deviations rather than long-term trend shifts. Consequently, the module primarily enhances short-horizon accuracy and robustness to bursty fluctuations, which is better reflected by MAE than by MSE.

Beyond aggregate error metrics, the Residual Correction Module plays a critical diagnostic role in preserving trajectory diversity and capturing idiosyncratic dynamics across similar products. Without this module, the model tends to generate highly similar trend forecasts for items belonging to the same fine-grained category, resulting in homogenized predictions that fail to reflect product-specific external influences. By explicitly modeling residual signals induced by latent exogenous factors, the proposed module reintroduces trajectory heterogeneity and substantially improves realism in multi-step rollouts.

\begin{figure}
    \centering
    \includegraphics[width=0.49\textwidth]{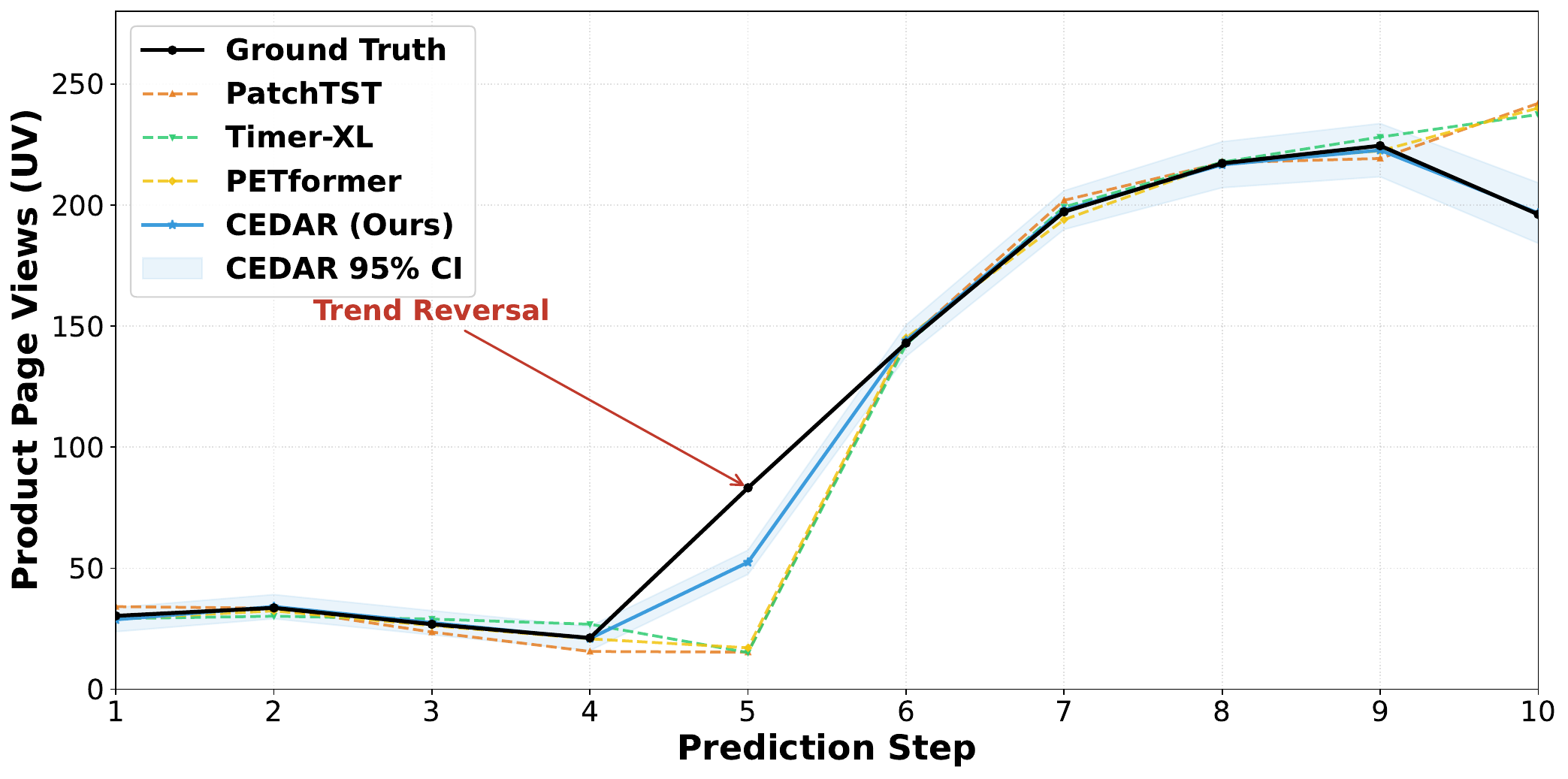}
    \caption{Visualization of predicted views trajectories with \textit{next-1} setting of different baselines. }
    \label{fig:injection}
\end{figure}

We further illustrate the critical effect of external shock modeling through a real-world case study shown in Figure~\ref{fig:injection}. The example corresponds to a sudden viral trend on social media surrounding ``Tanghulu with Naipizi'' (a fusion snack combining milk skin and candied hawthorn), which triggered a sharp surge in demand for a particular tanghulu supplier. In this setting, the model is tasked with performing \textit{next-1} forecasting, using observations from the preceding five weeks to predict the state vector of the subsequent week.

As observed, baseline models lacking explicit external event modeling mechanisms continue to extrapolate along previously observed trajectories, failing to anticipate the upcoming sales spike. In contrast, the Residual Correction Module successfully extracts highly relevant keywords (e.g., ``Tanghulu'') from trending search and external hotspot signals, enabling the model to correctly predict the abrupt increase in traffic and sales. This qualitative result demonstrates that the proposed module effectively captures latent external shocks and injects critical event-driven signals into the forecasting process.

Taken together, both quantitative and qualitative evidence confirms that the Residual Correction Module is essential for robust short-term forecasting under volatile market conditions, significantly enhancing the model’s responsiveness to bursty external events and improving the fidelity of simulated trajectories.

\subsection{Public Dataset Generalization}

To evaluate generalization beyond Alibaba 1688, we conduct additional experiments on the public Kaggle Store Sales dataset. Although this dataset contains retail demand series, promotion-related covariates, and calendar events, it differs from our setting because it is organized at the store-family level and lacks merchant actions with explicit budget-planning semantics. Thus, this experiment mainly tests whether the event-aware residual decomposition transfers to a public retail forecasting scenario, rather than fully reproducing our counterfactual budget-planning task.

For this public benchmark, we construct event inputs from holiday metadata and use a training protocol consistent with our Alibaba 1688 experiments. The results are reported in Table~\ref{tab:public_store_sales}. CEDAR consistently outperforms the representative time-series baselines across all metrics under the next-5 setting. Specifically, CEDAR reduces MSE from $0.6321$ to $0.5819$ compared with the strongest baseline PETFormer, and also achieves lower MAE and NMSE than both PatchTST and PETFormer. These results suggest that the proposed decomposition between base forecasting and event-driven residual correction is not specific to the Alibaba 1688 dataset, and can also provide benefits in public retail forecasting scenarios where external events influence future demand.

\begin{table}[t]
\centering
\caption{Performance comparison on the public Kaggle Store Sales dataset under the next-5 setting.}
\label{tab:public_store_sales}
\resizebox{0.7\columnwidth}{!}{
\begin{tabular}{lccc}
\toprule
\textbf{Metric} & \textbf{CEDAR} & \textbf{PatchTST} & \textbf{PETFormer} \\
\midrule
MSE $\downarrow$ & \textbf{0.5819} & 0.6814 & 0.6321 \\
MAE $\downarrow$ & \textbf{0.3680} & 0.4112 & 0.4623 \\
NMSE $\downarrow$ & \textbf{0.3778} & 0.4425 & 0.4104 \\
\bottomrule
\end{tabular}
}
\end{table}

\subsection{Online A/B Test Results (RQ4)}

To validate the practical effectiveness of CEDAR in real-world budget planning scenarios, we deploy the proposed framework in Alibaba 1688’s online advertising and marketing optimization system and conduct a large-scale A/B test from January 1 to January 30, 2026.
The initial deployment successfully engaged 239 cooperative merchants across 245 orders for the prediction service, facilitating a total transaction value of 8.77 million RMB with an initial repurchase rate of 61\%.

In the online setting, CEDAR is used to generate decision-conditioned sales simulations under alternative budget allocation strategies, which are then integrated into the platform’s recommendation and planning pipeline. The control group follows the existing production model based on a diffusion-based time series forecasting model with only total budgets, while the treatment group adopts CEDAR-driven budget planning and traffic allocation strategies.

The experimental results demonstrate substantial and consistent improvements. Specifically, merchants in the treatment group achieve a \textbf{13\%(46,471 vs 41,228) increase in lifetime value (LTV)} and a \textbf{15\% improvement in store-level return on investment (ROI)} on average. These gains are observed across both short-term promotional campaigns and long-term operational planning, indicating robust performance under diverse business conditions.

We attribute the observed improvements to two key factors. First, by explicitly modeling action-conditioned state transitions, CEDAR enables more accurate simulation of future demand trajectories under alternative marketing strategies, allowing merchants to proactively adjust budgets before traffic and conversion dynamics materialize. Second, the incorporation of external shock modeling further enhances responsiveness to bursty demand and emerging trends, enabling timely reallocation of marketing resources during critical periods. Together, these capabilities significantly reduce ineffective ad spend and improve the alignment between budget allocation and true market demand.

Overall, the online A/B test in Alibaba 1688 results provide strong empirical evidence that decision-conditioned simulation is not only theoretically well-motivated but also delivers tangible business value at scale. This validates the practical relevance of CEDAR for real-world merchant-facing decision support systems and highlights its potential for broader deployment in large-scale e-commerce platforms.

\section{Conclusion}
In this work, we propose \textbf{CEDAR}, a decision-conditioned sales simulation framework that explicitly decouples endogenous market dynamics from latent external shocks, while treating both merchant actions and product states as first-class modalities. CEDAR consists of two core components: an \emph{Action-Interleaved Transformer} that models action-conditioned state transitions through interleaved token sequences, and a \emph{Residual Correction Module} that leverages external event signals to predict and correct the discrepancy between base forecasts and real-world observations. To effectively optimize these two components, we adopt a two-stage training strategy that stabilizes long-horizon rollout and enhances robustness under volatile market conditions. Extensive experiments on the large-scale \textbf{E-Comm 15-Week} benchmark demonstrate consistent improvements over state-of-the-art baselines, while large-scale online A/B tests on the Alibaba 1688 platform further validate the practical value of CEDAR, yielding significant gains in merchant lifetime value and store-level return on investment.

In summary, this work highlights the importance of explicit action-conditioned modeling and external shock disentanglement for reliable what-if analysis and budget planning in e-commerce systems, paving the way for more robust and decision-aware forecasting frameworks in complex, intervention-driven environments.

\begin{acks}
This work was supported in part by the National Natural Science Foundation of China (Grant No. 62506348), the Natural Science Foundation of Anhui Province (Grant No. 2508085QF211), New Generation Artificial Intelligence-National Science and Technology Major Project (Grant No. 2025ZD0122601), the CCF-1688 Yuanbao Cooperation Fund (Grant No. CCF-Alibaba2025005), the National Key R\&D Program of China (Grant No. 2023YFF0725001), the National Natural Science Foundation of China (Grant No. 92370204), the Guangdong Basic and Applied Basic Research Foundation (Grant No. 2023B1515120057), the Key-Area Special Project of Guangdong Provincial Ordinary Universities (2024ZDZX1007).
\end{acks}

\balance
\bibliographystyle{ACM-Reference-Format}
\bibliography{ref}

@InProceedings{zhou2022fedformer,
  title = 	 {{FED}former: Frequency Enhanced Decomposed Transformer for Long-term Series Forecasting},
  author =       {Zhou, Tian and Ma, Ziqing and Wen, Qingsong and Wang, Xue and Sun, Liang and Jin, Rong},
  booktitle = 	 {Proceedings of the 39th International Conference on Machine Learning},
  pages = 	 {27268--27286},
  year = 	 {2022},
  editor = 	 {Chaudhuri, Kamalika and Jegelka, Stefanie and Song, Le and Szepesvari, Csaba and Niu, Gang and Sabato, Sivan},
  volume = 	 {162},
  series = 	 {Proceedings of Machine Learning Research},
  month = 	 {17--23 Jul},
  publisher =    {PMLR},
  url = 	 {https://proceedings.mlr.press/v162/zhou22g.html}
}

@inproceedings{wu2021autoformer,
 author = {Wu, Haixu and Xu, Jiehui and Wang, Jianmin and Long, Mingsheng},
 booktitle = {Advances in Neural Information Processing Systems},
 editor = {M. Ranzato and A. Beygelzimer and Y. Dauphin and P.S. Liang and J. Wortman Vaughan},
 pages = {22419--22430},
 publisher = {Curran Associates, Inc.},
 title = {Autoformer: Decomposition Transformers with Auto-Correlation for Long-Term Series Forecasting},
 url = {https://proceedings.neurips.cc/paper_files/paper/2021/file/bcc0d400288793e8bdcd7c19a8ac0c2b-Paper.pdf},
 volume = {34},
 year = {2021}
}

@article{zhou2021informer, title={Informer: Beyond Efficient Transformer for Long Sequence Time-Series Forecasting}, volume={35}, url={https://ojs.aaai.org/index.php/AAAI/article/view/17325}, DOI={10.1609/aaai.v35i12.17325}, abstractNote={Many real-world applications require the prediction of long sequence time-series, such as electricity consumption planning. Long sequence time-series forecasting (LSTF) demands a high prediction capacity of the model, which is the ability to capture precise long-range dependency coupling between output and input efficiently. Recent studies have shown the potential of Transformer to increase the prediction capacity. However, there are several severe issues with Transformer that prevent it from being directly applicable to LSTF, including quadratic time complexity, high memory usage, and inherent limitation of the encoder-decoder architecture. To address these issues, we design an efficient transformer-based model for LSTF, named Informer, with three distinctive characteristics: (i) a ProbSparse self-attention mechanism, which achieves O(L log L) in time complexity and memory usage, and has comparable performance on sequences’ dependency alignment. (ii) the self-attention distilling highlights dominating attention by halving cascading layer input, and efficiently handles extreme long input sequences. (iii) the generative style decoder, while conceptually simple, predicts the long time-series sequences at one forward operation rather than a step-by-step way, which drastically improves the inference speed of long-sequence predictions. Extensive experiments on four large-scale datasets demonstrate that Informer significantly outperforms existing methods and provides a new solution to the LSTF problem.}, number={12}, journal={Proceedings of the AAAI Conference on Artificial Intelligence}, author={Zhou, Haoyi and Zhang, Shanghang and Peng, Jieqi and Zhang, Shuai and Li, Jianxin and Xiong, Hui and Zhang, Wancai}, year={2021}, month={May}, pages={11106-11115} }

@misc{nie2023patchtst,
      title={A Time Series is Worth 64 Words: Long-term Forecasting with Transformers}, 
      author={Yuqi Nie and Nam H. Nguyen and Phanwadee Sinthong and Jayant Kalagnanam},
      year={2023},
      eprint={2211.14730},
      archivePrefix={arXiv},
      primaryClass={cs.LG},
      url={https://arxiv.org/abs/2211.14730}, 
}

@article{liu2023itransformer,
  title={itransformer: Inverted transformers are effective for time series forecasting},
  author={Liu, Yong and Hu, Tengge and Zhang, Haoran and Wu, Haixu and Wang, Shiyu and Ma, Lintao and Long, Mingsheng},
  journal={arXiv preprint arXiv:2310.06625},
  year={2023}
}

@inproceedings{
zheng2025valueguided,
title={Value-Guided Decision Transformer: A Unified Reinforcement Learning Framework for Online and Offline Settings},
author={Hongling Zheng and Li Shen and Yong Luo and Deheng Ye and Shuhan Xu and Bo Du and Jialie Shen and Dacheng Tao},
booktitle={The Thirty-ninth Annual Conference on Neural Information Processing Systems},
year={2025},
url={https://openreview.net/forum?id=Ogml5bxDH3}
}

@inproceedings{WenRobustPeriod20,
author = {Wen, Qingsong and He, Kai and Sun, Liang and Zhang, Yingying and Ke, Min and Xu, Huan},
title = {{RobustPeriod}: Time-Frequency Mining for Robust Multiple Periodicity Detection},
  booktitle={Proceedings of the 2021 International Conference on Management of Data (SIGMOD '21)},
  pages={205--215},
  year={2021}
}

@inproceedings{wen2020fastrobustSTL,
  title={Fast {RobustSTL}: Efficient and Robust Seasonal-Trend Decomposition for Time Series with Complex Patterns},
  author={Wen, Qingsong and Zhang, Zhe and Li, Yan and Sun, Liang},
  booktitle={Proceedings of the 26th ACM SIGKDD International Conference on Knowledge Discovery \& Data Mining (KDD '20)},
  pages={2203--2213},
  year={2020}
}

@inproceedings{wen2019robuststl,
  title={{RobustSTL}: A robust seasonal-trend decomposition algorithm for long time series},
  author={Wen, Qingsong and Gao, Jingkun and Song, Xiaomin and Sun, Liang and Xu, Huan and Zhu, Shenghuo},
  booktitle={Proceedings of the AAAI Conference on Artificial Intelligence},
  volume={33},
  number={01},
  pages={5409--5416},
  year={2019}
}

@article{cleveland1990stl,
  title={STL: A seasonal-trend decomposition},
  author={Cleveland, Robert B and Cleveland, William S and McRae, Jean E and Terpenning, Irma},
  journal={Journal of official statistics},
  volume={6},
  number={1},
  pages={3--73},
  year={1990}
}

@inproceedings{wang2018multilevel,
  title={Multilevel wavelet decomposition network for interpretable time series analysis},
  author={Wang, Jingyuan and Wang, Ze and Li, Jianfeng and Wu, Junjie},
  booktitle={Proceedings of the 24th ACM SIGKDD International Conference on Knowledge Discovery \& Data Mining},
  pages={2437--2446},
  year={2018}
}

@misc{chen2021decisiontransformer,
      title={Decision Transformer: Reinforcement Learning via Sequence Modeling}, 
      author={Lili Chen and Kevin Lu and Aravind Rajeswaran and Kimin Lee and Aditya Grover and Michael Laskin and Pieter Abbeel and Aravind Srinivas and Igor Mordatch},
      year={2021},
      eprint={2106.01345},
      archivePrefix={arXiv},
      primaryClass={cs.LG},
      url={https://arxiv.org/abs/2106.01345}, 
}

@article{janner2021offline,
  title={Offline reinforcement learning as one big sequence modeling problem},
  author={Janner, Michael and Li, Qiyang and Levine, Sergey},
  journal={Advances in neural information processing systems},
  volume={34},
  pages={1273--1286},
  year={2021}
}

@InProceedings{zheng2022online,
  title = 	 {Online Decision Transformer},
  author =       {Zheng, Qinqing and Zhang, Amy and Grover, Aditya},
  booktitle = 	 {Proceedings of the 39th International Conference on Machine Learning},
  pages = 	 {27042--27059},
  year = 	 {2022},
  editor = 	 {Chaudhuri, Kamalika and Jegelka, Stefanie and Song, Le and Szepesvari, Csaba and Niu, Gang and Sabato, Sivan},
  volume = 	 {162},
  series = 	 {Proceedings of Machine Learning Research},
  month = 	 {17--23 Jul},
  publisher =    {PMLR},
  url = 	 {https://proceedings.mlr.press/v162/zheng22c.html}
}

@inproceedings{yamagata2023q,
  title={Q-learning decision transformer: Leveraging dynamic programming for conditional sequence modelling in offline rl},
  author={Yamagata, Taku and Khalil, Ahmed and Santos-Rodriguez, Raul},
  booktitle={International Conference on Machine Learning},
  pages={38989--39007},
  year={2023},
  organization={PMLR}
}

@article{nakamoto2023cal,
  title={Cal-ql: Calibrated offline rl pre-training for efficient online fine-tuning},
  author={Nakamoto, Mitsuhiko and Zhai, Simon and Singh, Anikait and Sobol Mark, Max and Ma, Yi and Finn, Chelsea and Kumar, Aviral and Levine, Sergey},
  journal={Advances in Neural Information Processing Systems},
  volume={36},
  pages={62244--62269},
  year={2023}
}

@article{ansarichronos,
  title={Chronos: Learning the Language of Time Series},
  author={Ansari, Abdul Fatir and Stella, Lorenzo and Turkmen, Ali Caner and Zhang, Xiyuan and Mercado, Pedro and Shen, Huibin and Shchur, Oleksandr and Rangapuram, Syama Sundar and Arango, Sebastian Pineda and Kapoor, Shubham and others},
  journal={Transactions on Machine Learning Research}
}

@article{goswamimoment,
  title={MOMENT: A Family of Open Time-series Foundation Models},
  author={Goswami, Mononito and Szafer, Konrad and Choudhry, Arjun and Cai, Yifu and Li, Shuo and Dubrawski, Artur}
}

@article{wu2025doflow,
  title={DoFlow: Causal Generative Flows for Interventional and Counterfactual Time-Series Prediction},
  author={Wu, Dongze and Qiu, Feng and Xie, Yao},
  journal={arXiv e-prints},
  pages={arXiv--2511},
  year={2025}
}

@article{carta2018forecasting,
  title={Forecasting e-commerce products prices by combining an autoregressive integrated moving average (ARIMA) model and google trends data},
  author={Carta, Salvatore and Medda, Andrea and Pili, Alessio and Reforgiato Recupero, Diego and Saia, Roberto},
  journal={Future Internet},
  volume={11},
  number={1},
  pages={5},
  year={2018},
  publisher={MDPI}
}

@inproceedings{qiu-etal-2017-alime,
    title = "{A}li{M}e Chat: A Sequence to Sequence and Rerank based Chatbot Engine",
    author = "Qiu, Minghui  and
      Li, Feng-Lin  and
      Wang, Siyu  and
      Gao, Xing  and
      Chen, Yan  and
      Zhao, Weipeng  and
      Chen, Haiqing  and
      Huang, Jun  and
      Chu, Wei",
    editor = "Barzilay, Regina  and
      Kan, Min-Yen",
    booktitle = "Proceedings of the 55th Annual Meeting of the Association for Computational Linguistics (Volume 2: Short Papers)",
    month = jul,
    year = "2017",
    address = "Vancouver, Canada",
    publisher = "Association for Computational Linguistics",
    url = "https://aclanthology.org/P17-2079/",
    doi = "10.18653/v1/P17-2079",
    pages = "498--503"
}

@inproceedings{das2024decoder,
  title={A decoder-only foundation model for time-series forecasting},
  author={Das, Abhimanyu and Kong, Weihao and Sen, Rajat and Zhou, Yichen},
  booktitle={Forty-first International Conference on Machine Learning},
  year={2024}
}

@inproceedings{aksugift,
  title={GIFT-Eval: A Benchmark for General Time Series Forecasting Model Evaluation},
  author={Aksu, Taha and Woo, Gerald and Liu, Juncheng and Liu, Xu and Liu, Chenghao and Savarese, Silvio and Xiong, Caiming and Sahoo, Doyen},
  booktitle={NeurIPS Workshop on Time Series in the Age of Large Models}
}

@article{levine2020offline,
  title={Offline reinforcement learning: Tutorial, review, and perspectives on open problems},
  author={Levine, Sergey and Kumar, Aviral and Tucker, George and Fu, Justin},
  journal={arXiv preprint arXiv:2005.01643},
  year={2020}
}

@article{huang2024efficient,
  title={Efficient offline reinforcement learning with relaxed conservatism},
  author={Huang, Longyang and Dong, Botao and Zhang, Weidong},
  journal={IEEE Transactions on Pattern Analysis and Machine Intelligence},
  volume={46},
  number={8},
  pages={5260--5272},
  year={2024},
  publisher={IEEE}
}

@article{hu2024transforming,
  title={On transforming reinforcement learning with transformers: The development trajectory},
  author={Hu, Shengchao and Shen, Li and Zhang, Ya and Chen, Yixin and Tao, Dacheng},
  journal={IEEE Transactions on Pattern Analysis and Machine Intelligence},
  volume={46},
  number={12},
  pages={8580--8599},
  year={2024},
  publisher={IEEE}
}

@article{kumar2020conservative,
  title={Conservative q-learning for offline reinforcement learning},
  author={Kumar, Aviral and Zhou, Aurick and Tucker, George and Levine, Sergey},
  journal={Advances in neural information processing systems},
  volume={33},
  pages={1179--1191},
  year={2020}
}

@inproceedings{wilder2019melding,
  title={Melding the data-decisions pipeline: Decision-focused learning for combinatorial optimization},
  author={Wilder, Bryan and Dilkina, Bistra and Tambe, Milind},
  booktitle={Proceedings of the AAAI conference on artificial intelligence},
  volume={33},
  number={01},
  pages={1658--1665},
  year={2019}
}

@article{ye2024survey,
  title={A Survey of Time Series Foundation Models: Generalizing Time Series Representation with Large Language Model},
  author={Ye, Jiexia and Zhang, Weiqi and Yi, Ke and Yu, Yongzi and Li, Ziyue and Li, Jia and Tsung, Fugee},
  journal={CoRR},
  year={2024}
}

@article{nelson1998time,
  title={Time series analysis using autoregressive integrated moving average (ARIMA) models},
  author={Nelson, Brian K},
  journal={Academic emergency medicine},
  volume={5},
  number={7},
  pages={739--744},
  year={1998},
  publisher={Wiley Online Library}
}

@article{salinas2020deepar,
  title={DeepAR: Probabilistic forecasting with autoregressive recurrent networks},
  author={Salinas, David and Flunkert, Valentin and Gasthaus, Jan and Januschowski, Tim},
  journal={International journal of forecasting},
  volume={36},
  number={3},
  pages={1181--1191},
  year={2020},
  publisher={Elsevier}
}

@article{rafetseder2013counterfactual,
  title={Counterfactual reasoning: From childhood to adulthood},
  author={Rafetseder, Eva and Schwitalla, Maria and Perner, Josef},
  journal={Journal of experimental child psychology},
  volume={114},
  number={3},
  pages={389--404},
  year={2013},
  publisher={Elsevier}
}

@inproceedings{zhou2025balm,
  title={BALM-TSF: Balanced Multimodal Alignment for LLM-Based Time Series Forecasting},
  author={Zhou, Shiqiao and Sch{\"o}ner, Holger and Lyu, Huanbo and Fouch{\'e}, Edouard and Wang, Shuo},
  booktitle={Proceedings of the 34th ACM International Conference on Information and Knowledge Management},
  pages={4498--4508},
  year={2025}
}

@inproceedings{wolff2024spade,
  title={SPADE Split Peak Attention DEcomposition},
  author={Wolff, Malcolm and Olivares, Kin G and Oreshkin, Boris N and Ruan, Sunny and Yang, Sitan and Katoch, Abhinav and Ramasubramanian, Shankar and Zhang, Youxin and Mahoney, Michael W and Efimov, Dmitry and others},
  booktitle={NeurIPS Workshop on Time Series in the Age of Large Models},
  year={2024}
}

@article{potapczynski2024effectively,
  title={Effectively leveraging exogenous information across neural forecasters},
  author={Potapczynski, Andres and Olivares, Kin G and Wolff, Malcolm and Wilson, Andrew Gordon and Efimov, Dmitry and Quenneville-Belair, Vincent},
  year={2024}
}

@article{hu2025multi,
  title={Multi-Task Temporal Fusion Transformer for Joint Sales and Inventory Forecasting in Amazon E-Commerce Supply Chain},
  author={Hu, Zheqi and Hu, Yiwen and Li, Hanwu},
  journal={arXiv preprint arXiv:2512.00370},
  year={2025}
}

@article{punati2025temporal,
  title={Temporal Fusion Transformer for Multi-Horizon Probabilistic Forecasting of Weekly Retail Sales},
  author={Punati, Santhi Bharath and Kanta, Sandeep and Cheerala, Udaya Bhasker and Lanjewar, Madhusudan G and Damacharla, Praveen},
  journal={arXiv preprint arXiv:2511.00552},
  year={2025}
}

@article{woo2022etsformer,
  title={Etsformer: Exponential smoothing transformers for time-series forecasting},
  author={Woo, Gerald and Liu, Chenghao and Sahoo, Doyen and Kumar, Akshat and Hoi, Steven},
  journal={arXiv preprint arXiv:2202.01381},
  year={2022}
}

@article{oliveira2024evaluating,
  title={Evaluating the effectiveness of time series transformers for demand forecasting in retail},
  author={Oliveira, Jos{\'e} Manuel and Ramos, Patr{\'\i}cia},
  journal={Mathematics},
  volume={12},
  number={17},
  pages={2728},
  year={2024},
  publisher={MDPI AG}
}

@article{lin2024petformer,
  title={Petformer: Long-term time series forecasting via placeholder-enhanced transformer},
  author={Lin, Shengsheng and Lin, Weiwei and Wu, Wentai and Wang, Songbo and Wang, Yongxiang},
  journal={IEEE Transactions on Emerging Topics in Computational Intelligence},
  year={2024},
  publisher={IEEE}
}

@inproceedings{liu2025timer,
  title={Timer-XL: Long-Context Transformers for Unified Time Series Forecasting},
  author={Liu, Yong and Qin, Guo and Huang, Xiangdong and Wang, Jianmin and Long, Mingsheng},
  booktitle={The Thirteenth International Conference on Learning Representations}
}

@article{lim2021temporal,
  title={Temporal fusion transformers for interpretable multi-horizon time series forecasting},
  author={Lim, Bryan and Ar{\i}k, Sercan {\"O} and Loeff, Nicolas and Pfister, Tomas},
  journal={International journal of forecasting},
  volume={37},
  number={4},
  pages={1748--1764},
  year={2021},
  publisher={Elsevier}
}

@inproceedings{qin2025scihorizon,
  title={Scihorizon: Benchmarking ai-for-science readiness from scientific data to large language models},
  author={Qin, Chuan and Chen, Xin and Wang, Chengrui and Wu, Pengmin and Chen, Xi and Cheng, Yihang and Zhao, Jingyi and Xiao, Meng and Dong, Xiangchao and Long, Qingqing and others},
  booktitle={Proceedings of the 31st ACM SIGKDD Conference on Knowledge Discovery and Data Mining V. 2},
  pages={5754--5765},
  year={2025}
}

@article{wang2026face,
  title={Face: A general framework for mapping collaborative filtering embeddings into llm tokens},
  author={Wang, Chao and Song, Yixin and Ye, Jinhui and Qin, Chuan and Shen, Dazhong and Liu, Lingfeng and Wang, Xiang and Zhang, Yanyong},
  journal={Advances in Neural Information Processing Systems},
  volume={38},
  pages={146012--146039},
  year={2026}
}

@inproceedings{wang2021variable,
  title={Variable interval time sequence modeling for career trajectory prediction: Deep collaborative perspective},
  author={Wang, Chao and Zhu, Hengshu and Hao, Qiming and Xiao, Keli and Xiong, Hui},
  booktitle={Proceedings of the Web Conference 2021},
  pages={612--623},
  year={2021}
}

@article{wang2021personalized,
  title={Personalized and explainable employee training course recommendations: A bayesian variational approach},
  author={Wang, Chao and Zhu, Hengshu and Wang, Peng and Zhu, Chen and Zhang, Xi and Chen, Enhong and Xiong, Hui},
  journal={ACM Transactions on Information Systems (TOIS)},
  volume={40},
  number={4},
  pages={1--32},
  year={2021},
  publisher={ACM New York, NY}
}

@article{wang2023setrank,
  title={Setrank: A setwise bayesian approach for collaborative ranking in recommender system},
  author={Wang, Chao and Zhu, Hengshu and Zhu, Chen and Qin, Chuan and Chen, Enhong and Xiong, Hui},
  journal={ACM Transactions on Information Systems},
  volume={42},
  number={2},
  pages={1--32},
  year={2023},
  publisher={ACM New York, NY, USA}
}

@inproceedings{liu2026rethinking,
  title={Rethinking Popularity Bias in Collaborative Filtering via Analytical Vector Decomposition},
  author={Liu, Lingfeng and Song, Yixin and Shen, Dazhong and Yin, Bing and Li, Hao and Zhang, Yanyong and Wang, Chao},
  booktitle={Proceedings of the 32nd ACM SIGKDD Conference on Knowledge Discovery and Data Mining V. 1},
  pages={879--890},
  year={2026}
}

@inproceedings{10.1145/3774904.3792495,
author = {Zhang, Ranxu and Meng, Junjie and Sun, Ying and Xu, Ziqi and Yin, Bing and Li, Hao and Zhang, Yanyong and Wang, Chao},
title = {MCLMR: A Model-Agnostic Causal Learning Framework for Multi-Behavior Recommendation},
year = {2026},
isbn = {9798400723070},
publisher = {Association for Computing Machinery},
address = {New York, NY, USA},
url = {https://doi.org/10.1145/3774904.3792495},
doi = {10.1145/3774904.3792495},
booktitle = {Proceedings of the ACM Web Conference 2026},
pages = {6481–6492},
numpages = {12},
location = {United Arab Emirates},
series = {WWW '26}
}

@misc{meng2026turningsemanticstopologyllmdriven,
      title={Turning Semantics into Topology: LLM-Driven Attribute Augmentation for Collaborative Filtering}, 
      author={Junjie Meng and Ranxu zhang and Wei Wu and Rui Zhang and Chuan Qin and Qi Zhang and Qi Liu and Hui Xiong and Chao Wang},
      year={2026},
      eprint={2602.21099},
      archivePrefix={arXiv},
      primaryClass={cs.IR},
      url={https://arxiv.org/abs/2602.21099}, 
}

\appendix

\section{Details of LLM-based Hotspot Information Extraction}
\label{app.hostspot}

In this section, we detail the implementation of the Hotspot Information Extraction module. Raw social media trends are inherently noisy and often dominated by entertainment gossip, which can mislead forecasting models if used directly. To ensure high-quality semantic representations, we design a \textbf{two-stage LLM prompting pipeline}. The first stage filters noise and extracts commerce-relevant tags, while the second stage synthesizes these tags with calendar events into a single, coherent natural language sentence.

\subsection{Stage 1: Noise Filtering and Tag Extraction}
For each forecasting window $t$, we first collect raw trending lists $\mathcal{T}_t$ from major social platforms (eg.Red notebook and Douyin ) over the past week ($[t-7, t-1]$). In this stage, the LLM acts as an information filter. We prompt it to discard irrelevant news and extract only keywords (tags) indicative of potential product demand.

\begin{tcolorbox}[title=\textbf{Prompt 1: Commerce Tag Extraction}, colback=gray!5, colframe=black!70, arc=2mm]
\textbf{Role:} You are an e-commerce data analyst.

\textbf{Task:} Extract product-related keywords from the following noisy social media trending list.

\textbf{Input:} 
\begin{itemize}
    \item \textbf{Raw Trends:} \texttt{\{Raw\_Trend\_List\}}
\end{itemize}

\textbf{Instructions:}
1. Ignore all entertainment gossip, celebrity news, and non-commercial events.
2. Identify trends that could trigger consumer purchases (e.g., sudden weather changes, viral outdoor activities).
3. Extract concise product tags or categories related to these trends.

\textbf{Output Format:} A comma-separated list of tags (e.g., "camping tents, down jackets, traditional gifts").
\end{tcolorbox}

Let the output of this first stage be a set of filtered commerce tags, denoted as $\mathcal{K}_t$.

\subsection{Stage 2: Semantic Synthesis with Calendar Events}
Simply embedding a list of isolated tags $\mathcal{K}_t$ fails to capture the underlying market narrative and temporal context. Therefore, in the second stage, we fuse $\mathcal{K}_t$ with a structured list of upcoming calendar events $\mathcal{E}_t$ (e.g., public holidays, shopping festivals) occurring within the target window $t$. The LLM is instructed to synthesize these elements into \textbf{semantically complete sentences} $S_t$.

\begin{tcolorbox}[title=\textbf{Prompt 2: Narrative Synthesis}, colback=gray!5, colframe=black!70, arc=2mm]
\textbf{Role:} You are a senior market strategist.

\textbf{Task:} Synthesize the extracted product tags and upcoming holidays into a concise market summary.

\textbf{Inputs:}
\begin{itemize}
    \item \textbf{Extracted Product Tags:} \texttt{\{$\mathcal{K}_t$\}}
    \item \textbf{Upcoming Holidays/Events:} \texttt{\{$\mathcal{E}_t$\}}
\end{itemize}

\textbf{Instructions:}
1. First talks about the upcoming event and holidays.
2. Combine all information into \textbf{semantically complete sentences} that describes the overarching market demand drivers for the next week.

\textbf{Constraint:} You must output exactly follow instruction. Do not use bullet points or lists.

\textbf{Output Example:} 
"Driven by the upcoming Mid-Autumn Festival and the recent viral outdoor trend, the current market demand is highly concentrated on traditional gift boxes and premium camping gear."
\end{tcolorbox}

\subsection{Context Embedding Generation}
The output of Stage 2 ars fluent sentences $S_t$ that encapsulates the macro-environmental factors and market dynamics. Finally, we employ a pre-trained text encoder to transform $S_t$ into a dense vector representation:
$$
\mathbf{h}_t = \text{Encoder}(S_t) \in \mathbb{R}^{d_h}
$$
where $d_h$ is the embedding dimension. This dense hotspot embedding $\mathbf{h}_t$ serves as a critical global exogenous context signal for our framework, empowering the model to adjust its predictions based on a semantic understanding of concurrent external drivers.

\section{Causal Identifiability and Structural Invariance Analysis of CEDAR}

From the perspective of Structural Causal Models (SCM), the e-commerce sales system is formalized as a controlled stochastic process. We posit that the underlying data generation process follows an Additive Noise Model defined as:

\begin{equation}
    S_{t+1} := f_{\theta}(S_{\le t}, A_{t+1}) + \mathcal{E}(Z_{t+1}, U_t)
    \label{eq:scm_structural_en}
\end{equation}

where $S$ denotes endogenous states, $A$ represents merchant interventions, $Z$ represents observed exogenous events, and $U$ denotes unobserved noise. 

The CEDAR framework leverages this structural equation to achieve an effective approximation of the interventional distribution $P(S_{t+1} \mid do(A_{t+1}), S_{\le t})$ through architectural causal disentanglement. Specifically: (1) Learning Action-Conditioned Transition Operators: The AIT module in the first stage leverages an interleaved sequence structure to explicitly model the endogenous mechanism $f_{\theta}$. This approximation relies on the assumption of Sequential Ignorability, which posits that potential outcomes $S(a)$ are conditionally independent of the current action given the history $\mathcal{H}_t = \{S_{\le t}, Z_{\le t+1}\}$:
\begin{equation}
    S_{t+1}(a) \perp\!\!\!\perp A_{t+1} \mid \mathcal{H}_t, \quad \forall a \in \mathcal{A}
    \label{eq:sequential_ignorability_en}
\end{equation}
 (2) Orthogonal Decomposition of Exogenous Shocks: The residual correction module in the second stage explicitly captures the exogenous term $\mathcal{E}$. By incorporating LLM-enhanced event semantics $Z$, the model effectively blocks back-door paths induced by latent confounders such as market trends.

This separated modeling of Endogenous Mechanism + Exogenous Perturbation fundamentally ensures the \textbf{Structural Invariance} of the model during Counterfactual Simulation. It enables robust estimation of the marginal causal effects derived from hypothetical budget strategies, thereby overcoming the confounding bias inherent in traditional correlation-based models.

\section{Time Cost Analysis}
\label{app:time_cost}
The wall-clock cost analysis is reported in Table~\ref{tab:time_cost}. Compared with standard forecasting baselines, CEDAR introduces additional computation mainly from the two-stage training pipeline and the LLM-based event embedding construction. Specifically, Stage I takes about 154 minutes to train, while Stage II adds another 80 minutes for learning the residual correction module. The total training time of CEDAR is therefore approximately 234 minutes, which is higher than PatchTST but substantially lower than PETFormer in our implementation. The event embedding generation requires around 4 hours; however, this step is performed only once as offline preprocessing and the resulting embeddings can be reused across downstream training and inference. Therefore, its cost is amortized over all products, time windows, and subsequent model updates. Considering the consistent improvements in both short-horizon and long-horizon forecasting accuracy, the additional computational overhead is acceptable for large-scale industrial deployment, especially in budget planning scenarios where simulation quality is more critical than one-time preprocessing cost.

\begin{table}[t]
\centering
\caption{Wall-clock cost analysis of CEDAR and representative baselines. The LLM-based event embedding generation is a one-time offline preprocessing cost.}
\label{tab:time_cost}
\resizebox{\columnwidth}{!}{
\begin{tabular}{lcccl}
\toprule
\textbf{Component} & \textbf{Time / Epoch} & \textbf{Epochs} & \textbf{Total Time} & \textbf{Notes} \\
\midrule
CEDAR Stage I & $\approx$ 11 min & 14 & 154 min (2h 34m) & Base-stage training \\
CEDAR Stage II & $\approx$ 16 min & 5 & 80 min (1h 20m) & Residual correction training \\
Event embedding generation & -- & -- & $\approx$ 4h & One-time offline cost \\
PatchTST & $\approx$ 7 min & 17 & 119 min (1h 59m) & Baseline reference \\
PETFormer & $\approx$ 21 min & 31 & 651 min (10h 51m) & Baseline reference \\
\bottomrule
\end{tabular}
}
\end{table}

\section{Long-horizon Performance Comparison}

To further examine rollout stability, we extend the evaluation horizon from next-5 and next-10 to next-15, next-20, and next-25, using the same 10-week historical input. As shown in Table~\ref{tab:long_horizon}, all models exhibit larger errors as the horizon increases, which is expected due to autoregressive error accumulation. Nevertheless, CEDAR degrades more gracefully than the strongest baselines. At next-25, CEDAR obtains an MSE of 1.612, substantially lower than PatchTST 2.847 and PETFormer 2.561. This result suggests that explicitly modeling action-conditioned transitions and event-driven residuals improves not only short-horizon accuracy but also the fidelity of long-horizon counterfactual rollouts.

\label{app:Long-horizon}
\begin{table}[t]
\centering
\caption{Long-horizon rollout performance on Alibaba 1688 measured by MSE. All settings use 10 weeks of historical observations as input.}
\label{tab:long_horizon}
\resizebox{0.7\columnwidth}{!}{
\begin{tabular}{lccc}
\toprule
\textbf{Horizon} & \textbf{CEDAR} & \textbf{PatchTST} & \textbf{PETFormer} \\
\midrule
next 5  & \textbf{0.182} & 0.424 & 0.434 \\
next 10 & \textbf{0.414} & 0.849 & 0.743 \\
next 15 & \textbf{0.734} & 1.526 & 1.382 \\
next 20 & \textbf{1.103} & 2.173 & 1.945 \\
next 25 & \textbf{1.612} & 2.847 & 2.561 \\
\bottomrule
\end{tabular}
}
\end{table}
\end{document}